\documentclass{article}
\usepackage{collas2026_conference,times}
\usepackage{amsmath,amsfonts,bm}

\def\eqref#1{equation~\ref{#1}}
\def\1{\bm{1}}

\DeclareMathAlphabet{\mathsfit}{\encodingdefault}{\sfdefault}{m}{sl}
\SetMathAlphabet{\mathsfit}{bold}{\encodingdefault}{\sfdefault}{bx}{n}

\usepackage[T1]{fontenc}
\usepackage[utf8]{inputenc}
\usepackage{graphicx}
\graphicspath{{Figures/}{images/}{Sections/}}
\usepackage{float}
\usepackage{wrapfig}
\usepackage{caption}
\usepackage{subcaption}
\usepackage{tikz}
\usetikzlibrary{arrows.meta,positioning,shapes,calc,fit}
\usepackage{amsmath,amssymb,amsthm}
\usepackage{comment}
\usepackage{algorithm}
\usepackage{algpseudocode}
\usepackage{rotating}
\usepackage{placeins}

\usepackage{hyperref}
\hypersetup{
    colorlinks=true,
    linkcolor=red,
    filecolor=magenta,
    urlcolor=blue,
    citecolor=purple,
    pdftitle={Proactive Context-Forecasted Safety Constraints for Nonstationary Reinforcement Learning},
    pdfauthor={Tim Tomashevskiy},
    pdfpagemode=FullScreen,
}

\newcommand{\parencite}[1]{\citep{#1}}
\newcommand{\textcite}[1]{\citet{#1}}

\newtheorem{theorem}{Theorem}

\theoremstyle{definition}

\collasfinalcopy
\title{Proactive Context-Forecasted Safety Constraints for Nonstationary Reinforcement Learning}
\author{
Tim Tomashevskiy\\
Department of Computing and Software\\
McMaster University\\
Hamilton, Ontario, Canada\\
\texttt{tomashet@mcmaster.ca}
}

\begin{document}
\maketitle

\begin{abstract}
Ensuring safety in reinforcement learning under nonstationarity requires anticipating changes in risk before they lead to unsafe behavior. Existing approaches typically rely on safety constraints defined at design time or updated reactively during execution, assuming that such constraints remain valid over time. However, in nonstationary environments with evolving contexts and changing driving layouts, these assumptions may fail.

We propose a framework for proactive safety constraint generation based on context forecasting. The approach infers latent environmental context from observations, predicts its future evolution, and constructs safety constraints adapted to anticipated conditions. This enables the agent to proactively avoid unsafe regions instead of reacting only after safety violations occur.

We evaluate the method in driving environments with structured context variation. The experiments include a sweep over nonstationarity intensities and additional held-out driving layouts, including highway, intersection, and racetrack scenarios. Results show that proactive constraint generation substantially reduces collisions under both seen and out-of-training nonstationarity intensities and generally remains effective across held-out driving layouts while maintaining usable task performance.

These findings suggest that context-based constraint generation is a promising approach for safe reinforcement learning under nonstationarity.
\end{abstract}
\section{Introduction}

Reinforcement learning (RL) is a general framework for learning sequential decision-making policies by interacting with an environment \parencite{sutton2018reinforcement}. Over the past decades, RL algorithms have shown remarkable empirical success in a broad variety of applications, including robotics, control, and complex games. Nevertheless, the application of RL algorithms to real-world systems also poses significant safety issues.

In real-world applications, an RL agent is often required to meet strict safety constraints. Failure to meet these constraints can result in expensive system failures or undesired system behavior. For this reason, there is a growing interest in the research community in \emph{safe reinforcement learning}, where RL agents are required to meet safety constraints both during learning and at deployment time. Some existing approaches include constrained Markov decision processes, Lyapunov-based verification, and algorithms that ensure safe exploration with high probability \parencite{berkenkamp2017safe, wachi2020safe}. These methods enable agents to improve their policies while avoiding unsafe states and actions.

Most existing safe RL methods assume that the environment is stationary and that safety constraints are known in advance. However, in real-world systems, these assumptions are often violated. Operating conditions, external disturbances, and system dynamics may change over time, introducing various forms of \emph{non-stationarity} into the learning problem~\parencite{chandak2022reinforcement}. As a result, safety constraints specified at design time may become outdated, overly conservative, or inconsistent with the current environment.

The problem becomes even more challenging in continual or lifelong learning, where agents interact with changing environments over extended time horizons and continuously update their policies. In such settings, fixed safety constraints may become inconsistent with evolving environmental conditions and safety requirements. Developing safety constraints that adapt to environmental change remains a major challenge.

This problem brings us to a very important question: How should one formulate safety constraints during the design phase when reinforcement learning agents are operating in environments that have the potential to change over time?

To address this problem, this paper presents a goal-oriented framework for formulating constraints in the context of continual reinforcement learning. In this framework, instead of assuming that the constraint functions are known in advance, the framework formulates safety constraints based on higher-level safety objectives. These objectives determine how the constraints can be modified based on the additional information obtained by the agent about the environment.
%%%

This formulation is referred to as the FC-Goal framework, where FC-Goal denotes \emph{Future Constraint Goals}. The main idea is to formulate safety constraints based on design-time safety goals and adapt the constraints as new information about the environment becomes available. In this paper, FC-Goal is instantiated as a proactive context-forecasted safety-constraint framework: latent environmental context is extracted from observations, future context trajectories are predicted, and the predicted context is used to synthesize anticipatory safety constraints before unsafe behavior occurs.
%%%

%\paragraph{Contributions}
%%%%%%%
The contributions of this paper can be summarized as follows:
\begin{itemize}
    \item We formalize proactive safety-constraint generation for reinforcement learning under episodic nonstationarity, where constraints depend on evolving latent context.

    \item We introduce the FC-Goal framework, instantiated through context discovery, multi-step context forecasting, and anticipatory safety-constraint generation.

    \item We provide a conditional high-probability horizon-level safety analysis showing how calibrated context uncertainty, tail-safe constraint prediction, and MPC-style filtering can jointly control violation probability over a forecast horizon.

    \item We evaluate the method across nonstationarity intensities and held-out \textit{highway-env} layouts, showing substantial collision-rate reduction with usable task performance.
\end{itemize}
%%%%%%%%%%
The rest of this paper is organized as follows. Section~\ref{sec:related_work} discusses related work in safe reinforcement learning and non-stationary learning. Section~\ref{sec:background} explains the background of this work. Section~\ref{sec:problem} presents the problem formulation. Section~\ref{sec:method_overview} describes the proposed FC-Goal framework and its proactive context-forecasted safety-constraint instantiation. Section~\ref{sec:implementation} explains implementation details. Appendix~\ref {sec:theory_final} analyzes the theoretical properties of the method, and Section~\ref{sec:experiments} presents experimental evaluation.

\section{Related Work}
\label{sec:related_work}

Safe reinforcement learning has been studied through constrained optimization, risk-sensitive objectives, safe exploration, and robust control. Our work is most closely related to approaches that address uncertainty in safety constraints and adaptation under nonstationarity.

\paragraph{Safe reinforcement learning under fixed safety models.}
A common formulation models safety through constrained MDPs (CMDPs), solved using Lagrangian or primal--dual optimization \parencite{altman1999constrained, achiam2017constrained, chow2019lyapunov, tessler2019reward}. Related approaches use risk-sensitive criteria such as VaR and CVaR to control rare but catastrophic failures \parencite{tamar2015policy, chow2015risk, prashanth2014cvar}. Model-based safe exploration methods further account for epistemic uncertainty using confidence bounds or Gaussian processes \parencite{turchetta2016safe, berkenkamp2016safe, berkenkamp2017safe}, while robust MDPs address uncertainty through ambiguity sets and worst-case optimization \parencite{iyengar2005robust, nilim2005robust}. Although these approaches provide principled notions of safety, they typically assume that the safety model, cost structure, or uncertainty set can be specified in advance and remains valid over time.

\paragraph{Adaptation under nonstationarity.}
A separate line of work addresses nonstationarity through meta-learning, latent context inference, and online adaptation \parencite{finn2017model, nagabandi2018learning, rakelly2019efficient, zintgraf2020varibad}. These methods improve adaptation to changing environments, but focus primarily on performance recovery rather than proactive safety under context uncertainty.
%%%%
\paragraph{Safe RL under nonstationarity and context-aware adaptation.}
Recent work combines safety with nonstationary or task-varying RL using context inference, meta-learning, or online adaptation. CASRL is one of the closest context-aware safe RL methods: it infers latent environment context with a probabilistic model and evaluates safety using uncertainty-aware trajectory sampling under prior safety constraints and nonstationary disturbances~\parencite{chen2021context}. However, CASRL uses inferred context mainly for safe adaptation and planning in the current regime; it does not explicitly forecast the future evolution of context and convert that forecast into a horizon of future safety constraints. Meta-safe RL methods provide reward and constraint-violation guarantees across related CMDP tasks~\parencite{khattar2024cmdp}, while PEARL$^+$ improves pre-adaptation safety by regularizing the prior policy before the agent observes a new task~\parencite{wen2022improved}. Online safe meta-RL under Markov task transitions provides mechanisms for all-time safety and sample-efficient meta-updates~\parencite{yuan2025all}. Safe continual RL under nonstationarity has also been surveyed and organized by constraint formulation, safety mechanism, and adaptation speed, including reactive, quick, and proactive safety adaptation~\parencite{tomashevskiy2026safe}. Overall, this literature shows that context and task structure are important for safe adaptation, but existing approaches generally use inferred context or task-transition structure to adapt policies, update safety mechanisms, or provide violation guarantees for the current or newly encountered regime. In contrast, our framework explicitly forecasts latent context dynamics and uses the predicted context trajectory to synthesize anticipatory safety constraints before unsafe behavior occurs.
%%%%%
\paragraph{Positioning of our work.}
Our approach combines context-aware adaptation with proactive safety-constraint generation by treating safety constraints as \emph{context-dependent} and explicitly modeling \emph{context evolution} under episodic nonstationarity. Instead of assuming a fixed safety model, we infer latent context from observations, predict its future evolution, and generate safety constraints proactively before action selection. In this sense, the proposed framework differs from CMDP and risk-sensitive approaches that rely on static safety specifications, from safe exploration methods that focus on stationary uncertainty, and from meta-learning approaches that adapt performance or current-task safety without synthesizing safety constraints for anticipated future conditions.

\section{Background and Motivation}
\label{sec:background}

Safety in reinforcement learning is typically enforced through explicit constraint formulations. Constraints can be specified either as a \emph{safe set} (reach--avoid specification) or as a \emph{constraint function} with a threshold,
\begin{equation}
f_C(s,a)\le \sigma.
\label{eq:constraint_threshold}
\end{equation}
Safe sets are commonly used in reach--avoid and safe exploration settings \parencite{berkenkamp2017safe, hsu2021reachavoidrl}, while constraint functions are standard in CMDPs and practical safe RL \parencite{achiam2017constrained, garcia2015comprehensive}. In most existing approaches, these constraints are defined at design time and treated as fixed.

Constraint formulations differ in the strength of safety guarantees. \emph{Hard constraints} enforce safety at every time step, while \emph{expected (CMDP-style) constraints} bound cumulative cost in expectation \parencite{altman1999constrained, achiam2017constrained}. Intermediate formulations include \emph{chance constraints}, which limit violation probability, and \emph{risk-sensitive} criteria such as CVaR, which control tail risks \parencite{tamar2015policy, chow2015risk}. \emph{Robust constraints} further account for model uncertainty through worst-case guarantees \parencite{iyengar2005robust, nilim2005robust}. 

A key limitation of these approaches is the assumption that safety constraints remain valid after deployment. In nonstationary environments, this assumption breaks down. Constraint specifications may be incomplete at design time, and distribution shifts can invalidate learned safety relationships between states, actions, and constraint signals. As a result, fixed constraints may become either unsafe or overly conservative as conditions evolve.

These limitations motivate adaptive, context-dependent safety mechanisms that can update constraints in response to changing environments. In this work, we address this challenge by generating safety constraints proactively from predicted context, enabling the agent to anticipate and avoid unsafe regions under nonstationarity.
\section{Problem Formulation}
\label{sec:problem}

We consider an agent interacting with an environment over episodes $i=1,2,\dots$. Each episode corresponds to a stationary MDP instance $M_i$, but the environment is \emph{nonstationary across episodes}:
\[
M_1, M_2, \dots, M_i, M_{i+1},\dots,
\qquad M_i \neq M_{i+1}.
\]
We assume that each $M_i$ can be summarized by a latent context variable $z_i$ that captures stationary properties of the episode (e.g., traffic density, aggressiveness, observation noise). Thus, $z_i$ is constant within episode $i$, and changes between episodes. Given a context time series $z_1,\dots,z_i$, we assume the sequence has a transition pattern, so that $z_{i+1}$ is predictable from history.

\subsection{Goal-Based Safety Signal}
\label{sec:goal_based}

We assume the agent is given a high-level safety goal, while the exact low-level constraint function is unknown at design time. In driving, the natural safety goal is collision avoidance. We operationalize this goal through a clearance margin $d_t$, which measures distance to the nearest obstacle relative to a speed-dependent safe distance. Safety requires $d_t>0$, and the constraint function is defined as a prediction of future clearance.

This goal-based view provides a practical way to define safety without enumerating all low-level requirements explicitly. Instead, the algorithm learns how clearance depends on latent context and uses forecasts of context to update constraints proactively.

\section{Method Overview: Proactive Context-Forecasted Constraints}
\label{sec:method_overview}

This section describes the FC-Goal framework and its proactive context-forecasted safety-constraint instantiation. The method operationalizes the high-level idea of deriving safety constraints from evolving safety goals by using three stages: context discovery, context prediction, and constraint forecasting. We first summarize the full pipeline and then provide the implementation details in Section~\ref{sec:implementation}.
\subsection{High-Level Overview}
\label{sec:highlevel_upgrades}

The proactive context-forecasted instantiation of FC-Goal is designed for episodic nonstationarity in driving. Each episode $i$ is treated as a stationary MDP instance summarized by an episode-level latent context $z_i$, while nonstationarity occurs through changes in context between episodes. The pipeline consists of three tiers.
\begin{figure}[H]
    \centering
    \includegraphics[width=0.95\linewidth]{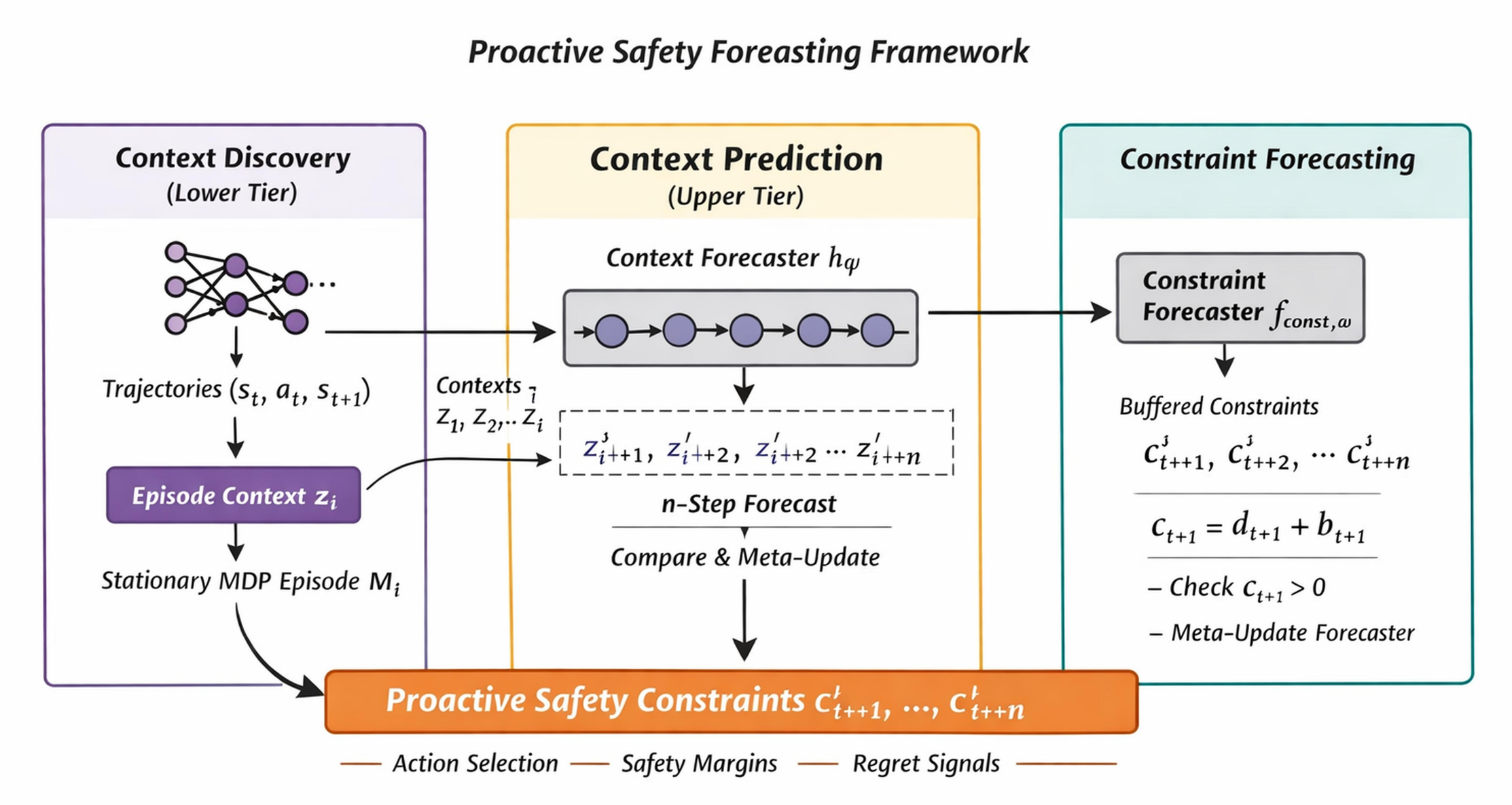}
    \caption{High-level pipeline of the proactive context-forecasted safety framework. Tier~1 extracts an episode context $\hat z_i$, Tier~2 forecasts future contexts and forms calibrated uncertainty sets, and Tier~3 predicts multi-step tail-safe clearance constraints. An MPC-style action filter enforces robust constraints over the forecast horizon, enabling control-aware proactive safety.}
    \label{fig:framework}
\end{figure}
\textbf{Tier~1: Context Discovery.}
The first tier infers an episode-level context embedding $\hat z_i$ from recent transition windows $(s_t,a_t,s_{t+1})$. This representation is intended to capture stationary episode properties such as traffic density, other-agent aggressiveness, observation noise, or other latent factors that affect safety and dynamics.

\textbf{Tier~2: Context Prediction.}
The second tier models the context time series $\{\hat z_1,\dots,\hat z_i\}$ and predicts an $n$-step future context trajectory for $n\ge 10$. The forecaster outputs a probabilistic prediction $(\mu_{i+1:i+n},\Sigma_{i+1:i+n})$. To account for context uncertainty, the forecast is calibrated using conformal prediction, yielding ellipsoidal uncertainty sets $\mathcal{Z}_{i+k}(\rho)$ around plausible future contexts.

\textbf{Tier~3: Constraint Forecasting and Enforcement.}
The third tier converts the predicted context trajectory into a horizon of safety constraints. In the driving setting, this is done by forecasting conservative tail estimates of future clearance, such as lower quantiles or CVaR-style margins, and forming proactive constraints $c'_{t+k}=\hat d^{(q)}_{t+k}-\epsilon$ for $k=1,\dots,n$. These predicted constraints are then enforced through a control-aware MPC-style safety filter. Candidate action sequences are rolled out with a fast ego model, and only actions satisfying robust multi-step constraints under the calibrated context uncertainty set are accepted. If no feasible action is found, the filter executes a conservative fallback action.

This pipeline turns context prediction into actionable safety decisions: rather than waiting for violations or adapting only after a shift is observed, the agent uses predicted context evolution to synthesize anticipatory safety constraints before unsafe behavior occurs.

\section{Implementation Details}
\label{sec:implementation}

This section describes an implementable instantiation of the proposed framework for driving under \emph{episodic nonstationarity}. We use $i$ to index episodes and $t$ to index time steps within an episode. Each episode $i$ corresponds to a stationary MDP instance characterized by an episode-level latent context $z_i$, while nonstationarity occurs only across episodes through changes in $z_i$. The method consists of three tiers: \textbf{Tier~1} (context extraction), \textbf{Tier~2} (multi-step context forecasting with conformal ellipsoids and meta-learning updates), and \textbf{Tier~3} (tail-safe constraint forecasting). Predicted constraints are enforced via a control-aware MPC-style safety filter. The prediction horizon is $n\ge 10$.

\subsection{Tier 1: Context Extraction via Representation Learning}

\paragraph{Inputs and outputs.}
Tier~1 receives short transition windows $\tau_i=\{(s_t,a_t,s_{t+1})\}_{t=0}^{K-1}$ collected at the beginning of episode $i$. Its output is an episode-level context embedding $\hat z_i\in\mathbb{R}^d$ (typically $d\in[8,32]$), which is treated as constant within the episode. In driving, $\hat z_i$ is intended to capture stationary episode properties such as traffic density, other-agent aggressiveness, and observation noise.

\paragraph{Architecture.}
For vector observations, we implement the context encoder $g_\phi$ as a GRU (or Transformer) over the transition sequence, followed by an MLP head producing $\hat z_i$. For image observations, we prepend a convolutional backbone to encode each $s_t$ before the sequence model. Optionally, the encoder outputs Gaussian parameters $(\mu^z_i,\Sigma^z_i)$, and we set $\hat z_i=\mu^z_i$.

\paragraph{Training objective.}
We train Tier~1 jointly with a context-conditioned dynamics model $p_\theta(s_{t+1}\mid s_t,a_t,\hat z_i)$ to ensure that the learned context is predictive of the episode dynamics. We use a next-state prediction loss (Gaussian NLL or MSE) and an episode-consistency regularizer:
\begin{equation}
\mathcal{L}_{\text{Tier1}}(\phi,\theta)
=
-\sum_{t}\log p_\theta(s_{t+1}\mid s_t,a_t,\hat z_i)
\;+\;
\lambda_{\text{cons}}\!\sum_{u,v\in i}\|\hat z_{i}^{(u)}-\hat z_{i}^{(v)}\|_2^2,
\label{eq:tier1_loss}
\end{equation}
where $\hat z_{i}^{(u)}$ and $\hat z_{i}^{(v)}$ are computed from different windows within the same episode. This encourages $\hat z_i$ to encode stationary MDP properties rather than transient state variation.

\subsection{Tier 2: Multi-step Context Forecasting with Regret Feedback}
Tier~2 models the time series of extracted contexts $\{\hat z_1,\dots,\hat z_i\}$ and predicts a multi-step forecast for future episode contexts. We implement the forecaster $h_\psi$ as a GRU/Transformer over the last $L$ contexts, producing Gaussian predictions
\[
(\mu_{i+1:i+n},\Sigma_{i+1:i+n})=h_\psi(\hat z_{i-L+1:i}),
\]
where $\Sigma$ is diagonal for efficiency. We train the forecaster using a discounted multi-step prediction loss:
\begin{equation}
\mathcal{L}^{(z)}(\psi)
=
\sum_{k=1}^{n} w_k \|\mu_{i+k}-\hat z_{i+k}\|_2^2,
\qquad
w_k=\gamma^{k-1}.
\label{eq:tier2_loss}
\end{equation}

\paragraph{Horizon prediction and regret.}
Tier~2 predicts contexts for $n\ge 10$ steps ahead. At the start of episode $i+1$, Tier~1 extracts the realized context $\hat z_{i+1}$, and Tier~2 computes a regret signal
\begin{equation}
r^{(z)}_{i+1}=\|\mu_{i+1}-\hat z_{i+1}\|.
\label{eq:tier2_regret}
\end{equation}
This regret is used to track forecasting performance and to drive online adaptation.

\paragraph{Off-policy meta-learning updates.}
To enable rapid adaptation to changing context-transition patterns, Tier~2 is updated using off-policy meta-learning on replayed episode subsequences stored in an episode buffer $\mathcal{D}=\{\hat z_i\}$. Each meta-task samples a short subsequence and constructs a support/query split: the support set emulates limited new evidence (e.g., one newly observed episode), while the query set evaluates multi-step forecasting performance. We apply a MAML/Reptile-style update to learn an initialization of $\psi$ that reduces regret after regime changes.

\subsection{Conformal Calibration: Ellipsoidal Context Uncertainty Sets}
Context forecasts are uncertain, and downstream safety evaluation should remain valid when the predicted context is slightly wrong. We therefore construct calibrated uncertainty sets around the Tier~2 forecast. For each calibration episode, we compute the Mahalanobis score
\begin{equation}
s_{i+1}=(\hat z_{i+1}-\mu_{i+1})^\top \Sigma_{i+1}^{-1}(\hat z_{i+1}-\mu_{i+1}),
\label{eq:mahal}
\end{equation}
and choose $\rho^2$ as the $(1-\alpha)$ quantile of scores stored in a calibration buffer $\mathcal{C}$. The resulting calibrated ellipsoid for episode $i+k$ is
\begin{equation}
\mathcal{Z}_{i+k}(\rho)=\left\{z:\,(z-\mu_{i+k})^\top \Sigma_{i+k}^{-1}(z-\mu_{i+k})\le \rho^2\right\}.
\label{eq:ellipsoid}
\end{equation}
In practice, we update $\rho$ using a sliding window of recent episodes to track slow drift while maintaining empirical coverage.

\subsection{Tier 3: Tail-safe Constraint Forecasting (Quantiles/CVaR)}
\paragraph{Driving safety signal.}
We define the safety objective as collision avoidance and operationalize it using a clearance margin relative to a speed-dependent safe distance:
\begin{equation}
d_{t} \;=\; \min_{j\in\mathcal{O}_t}\Big(\mathrm{dist}(\mathrm{ego},j)- (d_0+h v_t)\Big),
\label{eq:clearance}
\end{equation}
where $\mathcal{O}_t$ is the set of surrounding vehicles and $v_t$ is ego speed. A safety violation occurs when $d_t\le 0$.

\paragraph{Quantile forecasting.}
Tier~3 predicts a conservative lower-quantile of future clearance over horizon $n$:
\[
\hat d^{(q)}_{t+1:t+n}=f_\omega(x_t,\hat z_i),
\]
where $x_t$ includes ego kinematics, lane geometry features, and pooled obstacle features (e.g., attention-based set encoding). The quantile predictor is trained using the pinball loss:
\begin{equation}
\mathcal{L}_{\text{Tier3}}(\omega)
=
\sum_{k=1}^{n} w_k \,\rho_q\!\Big(d_{t+k}-\hat d^{(q)}_{t+k}\Big),
\qquad
\rho_q(u)=u\left(q-\mathbb{I}\{u<0\}\right).
\label{eq:pinball}
\end{equation}
We then define proactive constraints as
\begin{equation}
c'_{t+k}=\hat d^{(q)}_{t+k}-\epsilon,
\qquad k=1,\dots,n,
\label{eq:constraint}
\end{equation}
so that $c'_{t+k}>0$ enforces a tail-safe clearance margin with quantile level $q$ and safety margin $\epsilon$.

\paragraph{CVaR alternative.}
For higher conservatism, Tier~3 can be extended to a CVaR formulation by predicting a distribution (or samples) of $d_{t+k}$ and enforcing $\mathrm{CVaR}_\alpha(d_{t+k})>\epsilon$. In this work, quantile prediction provides a simple and effective tail-risk baseline.

\subsection{Control-aware Enforcement via MPC-style Safety Filtering}
Predicted constraints are enforced using an MPC-style safety filter. At each step $t$, we sample $M$ candidate control sequences $\mathbf{a}_{t:t+H-1}$ (steering/throttle/brake) and roll out a fast ego dynamics model (kinematic bicycle) to obtain predicted states $s'_{t+1:t+n}$. For each candidate, we evaluate robust multi-step safety under context uncertainty:
\begin{equation}
\min_{k\le n}\;\inf_{z\in\mathcal{Z}_{i+k}(\rho)}\left(\hat d^{(q)}(s'_{t+k},z)-\epsilon\right)\;\ge\;0.
\label{eq:mpc_constraint}
\end{equation}
The inner infimum is approximated using a small set of adversarial samples on the ellipsoid boundary. Among feasible candidates, we select the sequence maximizing progress and smoothness; if none is feasible, we execute a conservative fallback action (e.g., braking while maintaining lane).
%%%%%%%%%%%%%%%%%%%%%%%%%
\subsection{End-to-End Execution Loop}
\label{sec:impl_loop}

At a high level, each episode proceeds as follows: (i) Tier~1 extracts the episode-level context $\hat z_i$ from an initial transition window; (ii) Tier~2 forecasts future contexts and constructs conformal uncertainty sets; (iii) at each step, Tier~3 predicts tail-safe clearance constraints, and an MPC-style safety filter selects a feasible action under worst-case context realizations; (iv) data is stored and all tiers are periodically updated using replay buffers.

The full implementation pseudocode is provided in Appendix~\ref{app:impl_pseudocode}.

%%%%%%%%%%%%%%%%%%%%%%%%%%%%%%

\section{Experiments}
\label{sec:experiments}

We evaluate the proposed safety framework on \textit{merge-v0} from \textit{highway-env}. The main goal is to assess whether context-based safety constraints reduce unsafe behavior under stationary and increasingly nonstationary dynamics, while preserving task performance. The evaluation uses a sweep over switching frequencies, reports safety and task-performance metrics together, includes a compact component analysis, and provides full held-out-layout results in Appendix~\ref{app:heldout_envs}. We additionally evaluate \textit{highway-v0}, \textit{intersection-v0}, and \textit{racetrack-v0} as held-out stress-test layouts to assess whether the safety layer remains effective beyond the main \textit{merge-v0} benchmark.
%%%
\subsection{Setup and Metrics}

We train DQN and PPO agents with three random seeds on the main \textit{merge-v0} benchmark. Nonstationarity is controlled by $p_{\mathrm{stay}}$, the probability that the current environment context persists to the next episode; $p_{\mathrm{stay}}=1.0$ corresponds to the stationary setting, while smaller values indicate more frequent context changes. We evaluate $p_{\mathrm{stay}}\in\{1.0,0.95,0.85,0.70,0.50\}$. Unless otherwise stated, model selection and hyperparameter tuning use the stationary and milder switching regimes up to $p_{\mathrm{stay}}=0.85$, while $p_{\mathrm{stay}}=0.70$ and $p_{\mathrm{stay}}=0.50$ are treated as out-of-training nonstationarity intensities. For each algorithm and nonstationarity level, we compare the unconstrained agent (\emph{safety off}) with the full context-based safety mechanism (\emph{safety on}).

The primary metric is \emph{collision rate}. We also report \emph{final reward} as a task-performance metric and \emph{minimum distance} as an auxiliary proximity diagnostic. Minimum distance should not be interpreted as the sole safety criterion: a policy can reduce collisions while allowing smaller clearances in some maneuvers, reflecting a trade-off between strict collision avoidance, mobility, and flexibility.

%%%%%
\subsection{Results Across Nonstationarity Levels}

\begin{figure}[h]
\centering
\includegraphics[width=0.99\linewidth]{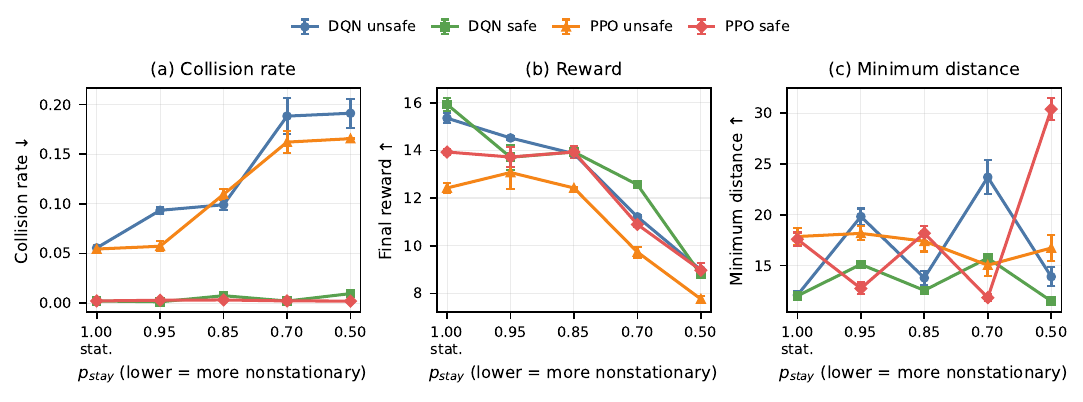}
\caption{Effect of increasing nonstationarity on the main \textit{merge-v0} benchmark. The horizontal axis shows $p_{\mathrm{stay}}$, where lower values indicate more frequent context switching. Collision rate is the primary safety metric: unconstrained agents become substantially more collision-prone as nonstationarity increases, while safety-enabled agents remain near zero. Error bars denote standard deviation across three seeds.}
\label{fig:nonstationarity_metrics}
\end{figure}

Figure~\ref{fig:nonstationarity_metrics} shows that the unconstrained baselines degrade as context switches become more frequent: DQN collision rate increases from $0.0555$ at $p_{\mathrm{stay}}=1.0$ to $0.1912$ at $p_{\mathrm{stay}}=0.50$, and PPO increases from $0.0544$ to $0.1656$. \mbox{In contrast,} the safety-enabled agents remain close to zero over the same sweep. At $p_{\mathrm{stay}}=0.70$, DQN drops from $0.1883$ to $0.0020$, and PPO drops from $0.1621$ to $0.0021$. Reward changes are modest relative to the collision-rate reduction, while minimum distance is mixed; we therefore interpret minimum distance as a diagnostic of proximity and mobility rather than as the primary safety target.

\begin{figure}[t]
\centering
\includegraphics[width=0.95\linewidth]{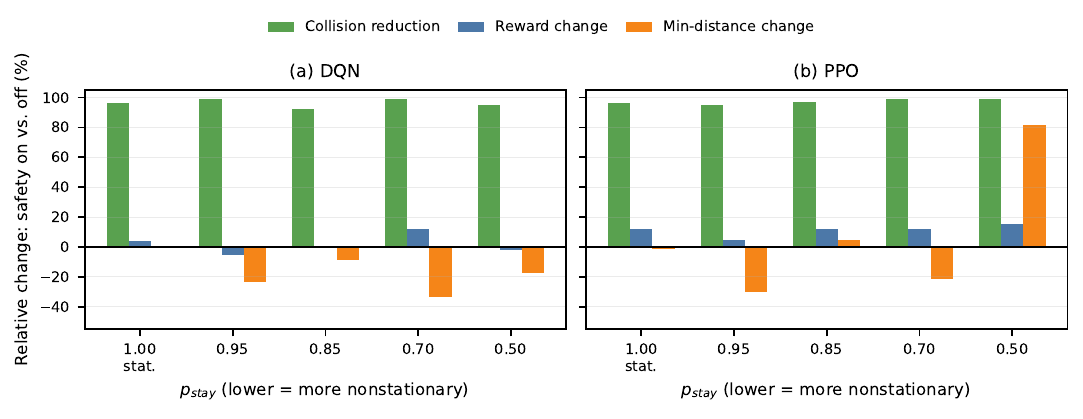}
\caption{Relative effect of enabling safety on the main \textit{merge-v0} benchmark. Collision-rate reduction is consistently large, while reward and minimum-distance changes are smaller and mixed, indicating a realistic safety--mobility trade-off rather than uniform improvement on all metrics.}
\label{fig:relative_effects}
\end{figure}

Figure~\ref{fig:relative_effects} summarizes the relative changes between safety-on and safety-off runs. Across all evaluated settings, enabling safety reduces collision rates by more than $90\%$ and usually by more than $97\%$. This supports the central empirical claim that context-based constraints primarily improve the target safety outcome while preserving usable task performance.

\begin{table}[t]
\centering
\small
\setlength{\tabcolsep}{4.5pt}
\renewcommand{\arraystretch}{1.12}
\begin{tabular}{lccc}
\hline
Variant & Collision rate $\downarrow$ & Final reward $\uparrow$ & Min. distance $\uparrow$ \\
\hline
No safety & $17.52\%$ & $10.47$ & $19.36$ \\
Fixed constraint & $6.72\%$ & $9.89$ & $27.52$ \\
Context-only constraint & $2.74\%$ & $10.91$ & $16.33$ \\
Forecasting without conformal calibration & $2.03\%$ & $10.51$ & $15.75$ \\
Full safety method & $0.21\%$ & $11.73$ & $13.78$ \\
\hline
\end{tabular}
\caption{Component analysis under strong nonstationarity on the main \textit{merge-v0} benchmark ($p_{\mathrm{stay}}=0.70$), aggregated across DQN and PPO. Collision rate is the primary safety metric; reward and minimum distance quantify mobility/proximity trade-offs. The full method achieves the lowest collision rate among all variants while maintaining usable reward; the minimum distance reflects the expected mobility/proximity trade-off.}
\label{tab:component_analysis}
\end{table}
%%%%%%%
\subsection{Held-out Layout Stress Test}
\label{subsec:heldout_layouts}

We further evaluate the safety layer on held-out \textit{highway-env} layouts: \textit{highway-v0}, \textit{intersection-v0}, and \textit{racetrack-v0}. These layouts test whether the same safety mechanism remains effective beyond the main \textit{merge-v0} benchmark and under different interaction geometries. Across the evaluated nonstationarity levels, the safety-enabled method reduces collision rates relative to the unconstrained baseline. Full per-layout results are reported in Appendix~\ref{app:heldout_envs}.
%%%%%%
\subsection{Discussion and Limitations}

The results strengthen the empirical evidence by evaluating multiple nonstationarity levels, reporting both DQN and PPO, separating collision reduction from auxiliary proximity diagnostics, adding held-out-layout stress tests, and including a compact component analysis. Figure~\ref{fig:nonstationarity_metrics} shows that decreasing $p_{\mathrm{stay}}$ increases collision rates for unconstrained agents, while the safety-enabled method remains effective even under out-of-training nonstationarity intensities. The held-out-layout stress test further suggests that the safety layer is not restricted to the main \textit{merge-v0} layout.

Table~\ref{tab:component_analysis} shows that fixed constraints already reduce collisions relative to no safety, but context-dependent and forecasted constraints reduce collisions further. This supports the claim that the benefit is not merely due to adding a static safety layer, but comes from adapting constraints to changing context. At the same time, reward and minimum distance show a realistic safety--mobility trade-off: the goal of the method is to prevent collisions rather than maximize clearance in every maneuver.

The evaluation remains limited in the range of nonstationarity mechanisms considered. Although the experiments vary the intensity of context switching through $p_{\mathrm{stay}}$, they do not systematically isolate which part of the context changes. In our formulation, visible and latent contexts jointly describe the state; future work should therefore distinguish shifts in visible context, shifts in latent context, and simultaneous shifts in both. These context changes may correspond to different statistical forms of distribution shift, such as covariate shift, label shift, or concept shift. Future work should study how each type of context shift affects the validity, conservativeness, and responsiveness of context-dependent constraints, especially when multiple shifts occur simultaneously.
\section{Conclusion}

This paper studied proactive safety constraint generation for reinforcement learning under episodic nonstationarity. The key idea is to treat safety constraints as context-dependent objects rather than fixed design-time specifications. By extracting latent context, forecasting its future evolution, and enforcing tail-safe clearance constraints through a control-aware safety filter, the proposed framework aims to prevent violations before they occur rather than only reacting after unsafe behavior is observed.

The experiments show that safety-enabled agents substantially reduce collision rates across a sweep of nonstationarity levels. As $p_{\mathrm{stay}}$ decreases, unsafe baselines generally become more collision-prone, while the safety-enabled runs maintain much lower collision rates for both DQN and PPO. The auxiliary reward and minimum-distance metrics indicate that this improvement is accompanied by a realistic safety--mobility trade-off rather than a uniform improvement on every metric. Additional held-out driving layouts, including highway, intersection, and racetrack scenarios, provide a stronger stress test than varying the switching intensity alone and support the more precise claim that the approach remains useful under both out-of-training nonstationarity intensities and held-out environment layouts.

These results support the central claim that context-dependent safety constraints are useful for RL systems operating in changing environments. At the same time, the current evaluation is not exhaustive. Future work should include larger-scale per-algorithm ablations, stronger reactive and fixed-constraint baselines, and more detailed diagnostics of context forecasting, calibration coverage, and intervention frequency.

\bibliographystyle{collas2026_conference}
\bibliography{bib}

\appendix
\section*{Appendix}
\vspace{-0.5em}

\section{Algorithm Details}
\label{app:impl_pseudocode}
\label{sec:appendix_algorithm}

\subsection{End-to-End Procedure}

\begin{algorithm}[h]
\caption{Proactive Context-Based Safety Constraint Generation}
\label{alg:final}
\begin{algorithmic}[1]
\Require Forecast horizon $n$, context encoder $g_\phi$, context forecaster $h_\psi$, constraint forecaster $f_\omega$, safety margin $\epsilon$, conformal radius $\rho$
\For{episode $i=1,2,\dots$}
    \State Collect an initial transition window $\tau_i=\{(s_t,a_t,s_{t+1})\}$
    \State Infer episode context $\hat z_i \leftarrow g_\phi(\tau_i)$
    \State Forecast future contexts $(\mu_{i+1:i+n},\Sigma_{i+1:i+n}) \leftarrow h_\psi(\hat z_{1:i})$
    \State Construct calibrated context uncertainty sets $\mathcal{Z}_{i+1:i+n}(\rho)$
    \For{step $t=0,\dots,T_i-1$}
        \State Observe current state $s_t$
        \State Predict tail-safe future clearances $\hat d^{(q)}_{t+1:t+n} \leftarrow f_\omega(s_t,\hat z_i)$
        \State Form proactive constraints $c'_{t+k}=\hat d^{(q)}_{t+k}-\epsilon$, for $k=1,\dots,n$
        \State Sample candidate action sequences $a_{t:t+H-1}^{(m)}$, $m=1,\dots,M$
        \State Roll out candidate sequences using a fast ego-dynamics model
        \State Keep only candidates satisfying
        \[
        \min_{k\le n}\inf_{z\in \mathcal{Z}_{i+k}(\rho)}
        \left(\hat d^{(q)}_{t+k}(z)-\epsilon\right) \ge 0
        \]
        \If{at least one feasible candidate exists}
            \State Execute the first action of the feasible sequence with best progress/smoothness score
        \Else
            \State Execute conservative fallback action
        \EndIf
        \State Store transition, realized clearance $d_{t+1}$, and violation indicator
    \EndFor
    \State Extract realized next context $\hat z_{i+1}$ when available
    \State Update context forecaster using prediction error $\|\mu_{i+1}-\hat z_{i+1}\|$
    \State Update constraint forecaster using clearance prediction error
    \State Update conformal calibration buffer and radius $\rho$
\EndFor
\end{algorithmic}
\end{algorithm}

\subsection{Constraint Formulation}

We define buffered constraints as:
\begin{equation}
c_{t+1} = d_{t+1} + b_{t+1},
\end{equation}
where $b_{t+1}\leq 0$ is a signed conservative correction accounting for uncertainty due to prediction error, context shift, and epistemic uncertainty.

A generic formulation is:
\begin{equation}
b_{t+1}
=
-B(\mathcal{U}_{d}, \mathcal{U}_{z}, \mathcal{U}_{\text{epi}}, \mathcal{V}),
\qquad B(\cdot)\geq 0,
\end{equation}
where these terms represent distance uncertainty, context uncertainty, epistemic uncertainty, and violation history, respectively. Thus, larger uncertainty produces a more negative correction and therefore a more conservative buffered clearance constraint.

\subsection{Learning Signals}

We define:
\begin{align}
r^{(z)}_{i+1} &= \|\mu_{i+1} - \hat z_{i+1}\|, \\
r^{(c)}_{t+1} &= |c'_{t+1} - c_{t+1}|.
\end{align}

These signals are used to update context forecasting and constraint prediction models.

\section{Additional Experimental Diagnostics}
\label{app:additional_experiments}

\begin{figure}[!t]
\centering
\includegraphics[width=0.68\linewidth]{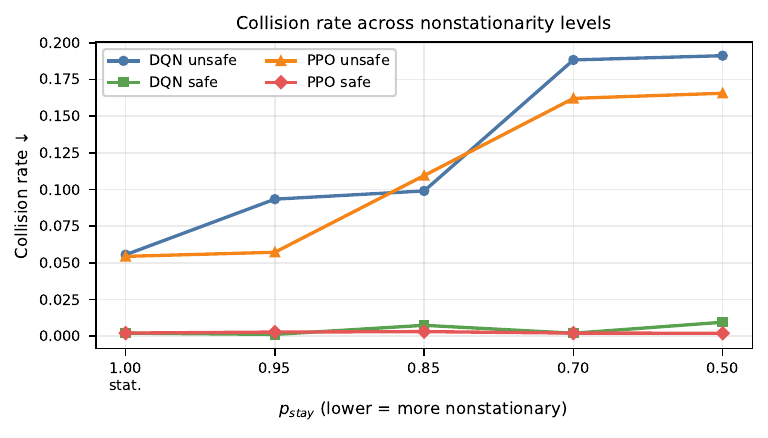}
\caption{Collision-rate sweep across the full $p_{\mathrm{stay}}$ range on the main \textit{merge-v0} benchmark. This figure reports the primary safety metric and shows that safety-enabled runs remain close to zero, while unconstrained baselines become more collision-prone as context switching becomes more frequent.}
\label{fig:app_collision_sweep}
\end{figure}

\begin{figure}[!t]
\centering
\includegraphics[width=0.68\linewidth]{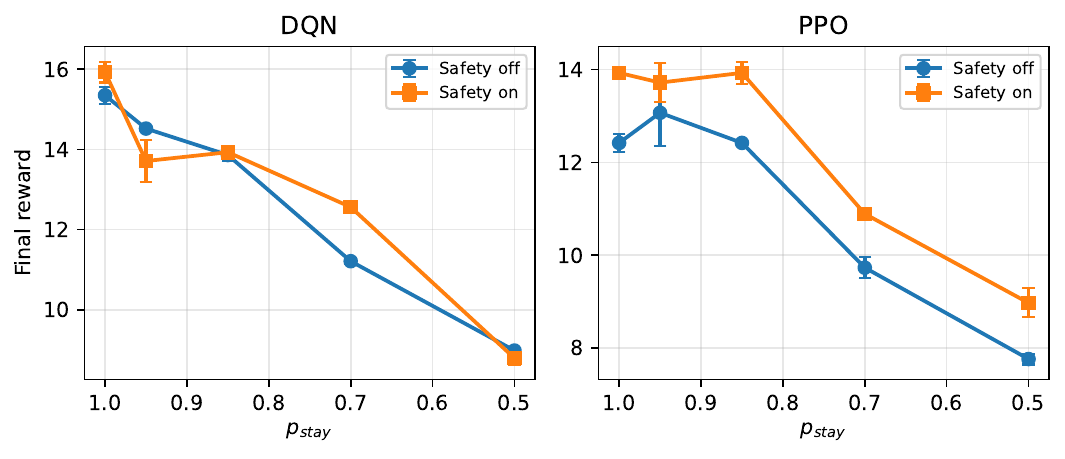}
\caption{Final-reward sweep across the full $p_{\mathrm{stay}}$ range. This figure reports task performance and shows that the safety layer preserves usable reward despite large collision-rate reductions, with some degradation under stronger nonstationarity.}
\label{fig:app_reward_sweep}
\end{figure}

\begin{figure}[!t]
\centering
\includegraphics[width=0.68\linewidth]{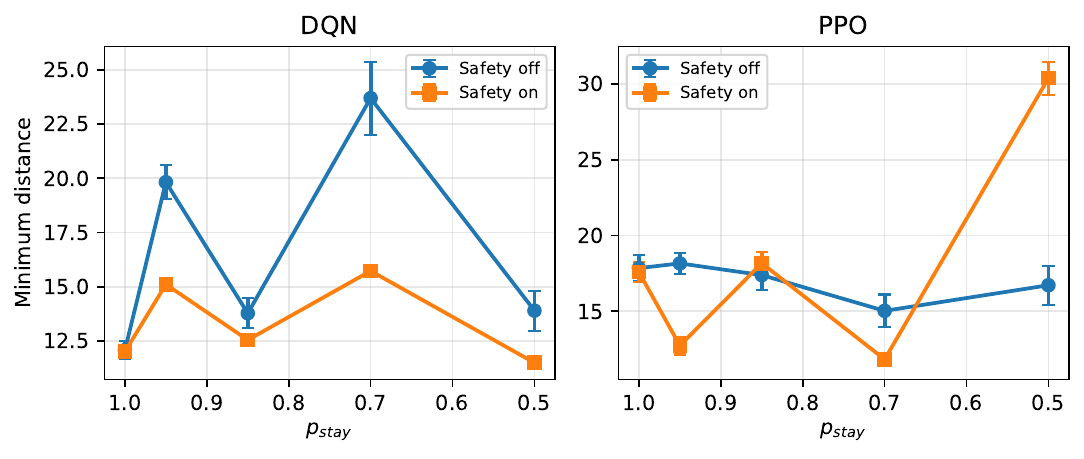}
\caption{Minimum-distance sweep across the full $p_{\mathrm{stay}}$ range. This figure reports an auxiliary proximity diagnostic rather than the primary safety target, illustrating that collision reduction is accompanied by realistic mobility/proximity trade-offs.}
\label{fig:app_distance_sweep}
\end{figure}

The main paper emphasizes compact visual summaries. Table~\ref{tab:main_results_full} provides the full numerical results used to generate Figures~\ref{fig:nonstationarity_metrics} and~\ref{fig:relative_effects}. These values are included in the appendix to keep the main experimental section focused. Figures~\ref{fig:app_collision_sweep}, \ref{fig:app_reward_sweep}, and~\ref{fig:app_distance_sweep} split the appendix metric sweep into collision rate, final reward, and minimum distance, respectively.

\begin{table}[H]
\centering
\scriptsize
\setlength{\tabcolsep}{3.2pt}
\renewcommand{\arraystretch}{1.08}
\resizebox{\linewidth}{!}{%
\begin{tabular}{llccccc}
\hline
Setting & Algo & Safety & Collision rate (\%) $\downarrow$ & Minimum distance $\uparrow$ & Final reward $\uparrow$ \\
\hline
Stationary ($p_{\mathrm{stay}}=1.00$) & DQN & off & $5.55 \pm 0.280$ & $12.08 \pm 0.42$ & $15.35 \pm 0.208$ \\
Stationary ($p_{\mathrm{stay}}=1.00$) & DQN & on  & $0.21 \pm 0.005$ & $12.01 \pm 0.06$ & $15.93 \pm 0.260$ \\
Stationary ($p_{\mathrm{stay}}=1.00$) & PPO & off & $5.44 \pm 0.430$ & $17.85 \pm 0.84$ & $12.42 \pm 0.191$ \\
Stationary ($p_{\mathrm{stay}}=1.00$) & PPO & on  & $0.21 \pm 0.004$ & $17.58 \pm 0.67$ & $13.93 \pm 0.121$ \\
\hline
Nonstationary ($p_{\mathrm{stay}}=0.95$) & DQN & off & $9.34 \pm 0.370$ & $19.82 \pm 0.79$ & $14.52 \pm 0.069$ \\
Nonstationary ($p_{\mathrm{stay}}=0.95$) & DQN & on  & $0.12 \pm 0.003$ & $15.11 \pm 0.08$ & $13.71 \pm 0.520$ \\
Nonstationary ($p_{\mathrm{stay}}=0.95$) & PPO & off & $5.72 \pm 0.510$ & $18.16 \pm 0.69$ & $13.07 \pm 0.710$ \\
Nonstationary ($p_{\mathrm{stay}}=0.95$) & PPO & on  & $0.28 \pm 0.010$ & $12.74 \pm 0.59$ & $13.72 \pm 0.416$ \\
\hline
Nonstationary ($p_{\mathrm{stay}}=0.85$) & DQN & off & $9.90 \pm 0.500$ & $13.78 \pm 0.69$ & $13.86 \pm 0.156$ \\
Nonstationary ($p_{\mathrm{stay}}=0.85$) & DQN & on  & $0.74 \pm 0.005$ & $12.55 \pm 0.09$ & $13.93 \pm 0.069$ \\
Nonstationary ($p_{\mathrm{stay}}=0.85$) & PPO & off & $10.95 \pm 0.560$ & $17.39 \pm 1.01$ & $12.42 \pm 0.042$ \\
Nonstationary ($p_{\mathrm{stay}}=0.85$) & PPO & on  & $0.31 \pm 0.003$ & $18.17 \pm 0.76$ & $13.93 \pm 0.242$ \\
\hline
Nonstationary ($p_{\mathrm{stay}}=0.70$) & DQN & off & $18.83 \pm 1.810$ & $23.69 \pm 1.68$ & $11.21 \pm 0.069$ \\
Nonstationary ($p_{\mathrm{stay}}=0.70$) & DQN & on  & $0.20 \pm 0.006$ & $15.73 \pm 0.14$ & $12.56 \pm 0.104$ \\
Nonstationary ($p_{\mathrm{stay}}=0.70$) & PPO & off & $16.21 \pm 1.130$ & $15.03 \pm 1.08$ & $9.73 \pm 0.225$ \\
Nonstationary ($p_{\mathrm{stay}}=0.70$) & PPO & on  & $0.21 \pm 0.003$ & $11.83 \pm 0.34$ & $10.89 \pm 0.104$ \\
\hline
Nonstationary ($p_{\mathrm{stay}}=0.50$) & DQN & off & $19.12 \pm 1.470$ & $13.90 \pm 0.92$ & $8.98 \pm 0.066$ \\
Nonstationary ($p_{\mathrm{stay}}=0.50$) & DQN & on  & $0.95 \pm 0.008$ & $11.50 \pm 0.08$ & $8.79 \pm 0.158$ \\
Nonstationary ($p_{\mathrm{stay}}=0.50$) & PPO & off & $16.56 \pm 0.070$ & $16.72 \pm 1.29$ & $7.76 \pm 0.118$ \\
Nonstationary ($p_{\mathrm{stay}}=0.50$) & PPO & on  & $0.20 \pm 0.002$ & $30.38 \pm 1.09$ & $8.97 \pm 0.312$ \\
\hline
\end{tabular}%
}
\caption{Training summary across three random seeds. Values are reported as mean $\pm$ standard deviation. Collision rate is reported as a percentage. Lower collision rate indicates better safety, higher final reward indicates better task performance, and minimum distance is reported as an auxiliary proximity diagnostic rather than the primary safety metric.}
\label{tab:main_results_full}
\end{table}

\FloatBarrier

\section{Held-out Layout Results}
\label{app:heldout_envs}

This appendix reports the held-out-layout stress test used to evaluate whether the safety layer remains effective beyond the main \textit{merge-v0} benchmark. The held-out environments are three \textit{highway-env} layouts: \textit{highway-v0}, \textit{intersection-v0}, and \textit{racetrack-v0}. These layouts introduce different road geometries and interaction patterns from the main merge setting.

\begin{figure}[h]
\centering
\includegraphics[width=0.82\linewidth]{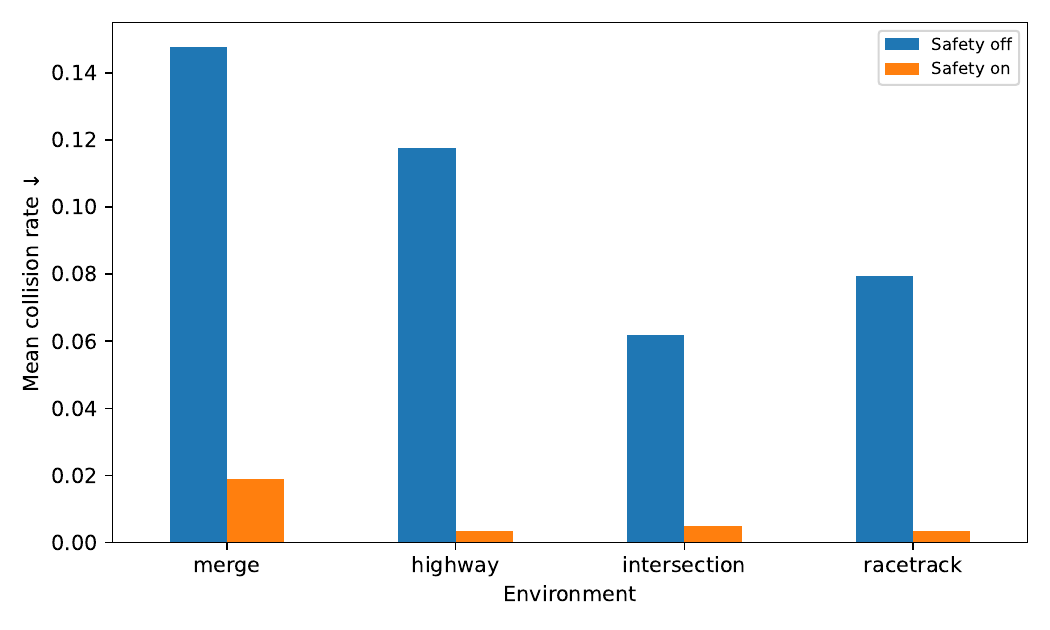}
\caption{Held-out layout stress test on \textit{highway-v0}, \textit{intersection-v0}, and \textit{racetrack-v0}. Bars show mean collision rate across DQN/PPO and evaluated nonstationarity levels. The safety-enabled method reduces collisions in each evaluated held-out layout.}
\label{fig:heldout_environment_collision}
\end{figure}

Table~\ref{tab:heldout_env_summary} reports aggregate collision rates and relative reductions for each held-out layout, while Figure~\ref{fig:heldout_environment_collision} visualizes the corresponding safety-off and safety-on collision rates. Across all three held-out layouts, the safety-enabled method reduces collision rates relative to the unconstrained baseline, suggesting that the safety layer is not restricted to the main \textit{merge-v0} environment.

\begin{table}[H]
\centering
\small
\setlength{\tabcolsep}{5pt}
\begin{tabular}{lccc}
\hline
Environment & Safety off (\%) & Safety on (\%) & Reduction \\
\hline
\texttt{merge} & 14.76\% & 1.88\% & 87.2\% \\
\texttt{highway} & 11.76\% & 0.34\% & 97.1\% \\
\texttt{intersection} & 6.18\% & 0.49\% & 92.0\% \\
\texttt{racetrack} & 7.94\% & 0.35\% & 95.7\% \\
\hline
\end{tabular}
\caption{Main and held-out layout collision summary averaged over DQN/PPO and all evaluated $p_{\mathrm{stay}}$ values. Collision rates are reported as percentages. The \textit{merge-v0} row corresponds to the main benchmark, while \textit{highway-v0}, \textit{intersection-v0}, and \textit{racetrack-v0} are held-out stress-test layouts. Safety-enabled runs reduce collisions across all layouts; the magnitude varies by road topology, so these results are presented as a stress test rather than a claim of universal transfer.}
\label{tab:heldout_env_summary}
\end{table}

\section{Theoretical Safety Guarantees}
\label{sec:theory_final}

Because the proposed method relies on learned context extraction, context forecasting, and tail-risk constraint prediction, its guarantees are conditional on calibration and predictive coverage assumptions. We therefore provide high-probability, horizon-level guarantees rather than unconditional hard-safety guarantees. The analysis focuses on multi-step proactive safety over a forecast horizon $n \ge 10$, using calibrated context uncertainty sets and tail-safe clearance prediction.

\subsection{Setup}
We assume episodic nonstationarity: each episode $i$ corresponds to a stationary MDP instance with latent context $z_i\in\mathbb{R}^d$, while $z_i$ changes only between episodes. Within an episode, at each step $t$ the agent observes a clearance margin
\begin{equation}
d_t \;=\; \min_{j\in\mathcal{O}_t}\Big(\mathrm{dist}(\mathrm{ego},j)-(d_0+h v_t)\Big),
\label{eq:clearance_theory}
\end{equation}
and safety requires $d_t>0$. The algorithm predicts an $n$-step tail-safe clearance forecast using Tier~3 and enforces a proactive constraint
\begin{equation}
c'_{t+k} \;=\; \hat d^{(q)}_{t+k}-\epsilon,
\qquad k=1,\dots,n,
\label{eq:constraint_def_theory}
\end{equation}
where $\hat d^{(q)}_{t+k}$ is a predicted $q$-quantile of clearance and $\epsilon>0$ is a fixed safety margin. Context uncertainty is handled via a conformal ellipsoid $\mathcal{Z}_{i}(\rho)$ derived from Tier~2.

\subsection{Assumptions}
We state the minimal conditions needed for high-probability safety.

\paragraph{A1 (Conformal context coverage).}
Tier~2 outputs Gaussian forecasts $(\mu_i,\Sigma_i)$ and a conformal radius $\rho$ is chosen such that the ellipsoid
\[
\mathcal{Z}_{i}(\rho)=\left\{z:\,(z-\mu_{i})^\top \Sigma_{i}^{-1}(z-\mu_{i})\le \rho^2\right\}
\]
satisfies the coverage property
\begin{equation}
\mathbb{P}(z_i\in \mathcal{Z}_{i}(\rho)) \;\ge\; 1-\alpha.
\label{eq:context_coverage}
\end{equation}

\paragraph{A2 (Robust quantile coverage).}
Tier~3 produces a context-conditioned quantile predictor $\hat d^{(q)}(s,z)$ such that for each $k\le n$,
\begin{equation}
\mathbb{P}\!\left(d_{t+k}\ge \hat d^{(q)}_{t+k}(z)\mid s_t,\pi\right)\;\ge\; 1-q,
\qquad \forall z\in \mathcal{Z}_i(\rho),
\label{eq:quantile_coverage}
\end{equation}
where the probability is over environment stochasticity and (optionally) model randomness. This assumption corresponds to a calibrated lower-quantile predictor, which can be approached in practice via quantile regression plus empirical calibration.

\paragraph{A3 (MPC feasibility).}
At each step, the MPC filter selects an action sequence whose predicted trajectory satisfies the robust constraint condition:
\begin{equation}
\min_{k\le n}\;\inf_{z\in\mathcal{Z}_i(\rho)}\left(\hat d^{(q)}_{t+k}(z)-\epsilon\right)\;\ge\;0.
\label{eq:mpc_feasible}
\end{equation}
If no feasible candidate exists, a conservative fallback is applied.

\subsection{Multi-step High-Probability Safety}
We now state a horizon-level safety guarantee for the executed trajectory.

\begin{theorem}[Proactive $n$-step safety under calibrated context and quantile prediction]
\label{thm:nstep_safety}
Assume A1--A3. Suppose the MPC filter enforces \eqref{eq:mpc_feasible} at time $t$. Then, with probability at least
\begin{equation}
1-\alpha-nq,
\label{eq:prob_bound}
\end{equation}
the realized clearance satisfies
\begin{equation}
d_{t+k}\geq\epsilon
\qquad \text{for all } k=1,\dots,n.
\label{eq:real_safety}
\end{equation}
\end{theorem}

\paragraph{Proof sketch.}
By A1, with probability at least $1-\alpha$ the true episode context satisfies $z_i\in\mathcal{Z}_i(\rho)$. Conditioned on this event, the robust constraint enforcement \eqref{eq:mpc_feasible} implies $\hat d^{(q)}_{t+k}(z_i)\ge \epsilon$ for all $k\le n$. By A2, for each fixed $k$, the probability that the realized clearance violates this bound is at most $q$, i.e., $\mathbb{P}(d_{t+k}<\hat d^{(q)}_{t+k}(z_i))\le q$. Applying a union bound across $k=1,\dots,n$ yields
\[
\mathbb{P}\left(\exists k\le n:\; d_{t+k}<\epsilon\right)\le \alpha + nq,
\]
which proves the claim. \qed

\subsection{Interpretation: What Type of Guarantees Are Provided?}
Theorem~\ref{thm:nstep_safety} provides a \emph{probabilistic (chance-style)} guarantee over the next $n$ steps. Importantly, the guarantee is \emph{proactive} and \emph{control-aware}: it applies to the trajectory induced by the selected MPC-filtered actions rather than to passive rollouts. The guarantee is not an unconditional hard constraint in the classical control-theoretic sense, since it depends on calibrated coverage properties of learned predictors. However, the failure probability is explicitly controlled by $(\alpha,q,n)$: decreasing $\alpha$ enlarges the context ellipsoid, decreasing $q$ increases conservatism in the quantile predictor, and increasing $n$ makes the guarantee stricter but more challenging.

\subsection{From Per-step Guarantees to Cumulative Safety}
Although the algorithm enforces a per-step margin, it also implies a cumulative bound on the probability of any violation over a time horizon $T$:
\[
\mathbb{P}(\exists t\le T:\; d_t<\epsilon)\le \left\lceil\frac{T}{n}\right\rceil(\alpha+nq),
\]
by applying Theorem~\ref{thm:nstep_safety} over consecutive blocks of length $n$ and union bounding the failure events. Thus, per-step proactive constraints yield a cumulative risk control interpretation, where the total violation probability grows at most linearly with time.

\subsection{Discussion}
The guarantee above highlights the role of the three upgrades: conformal calibration provides explicit context uncertainty coverage, quantile forecasting controls tail risk in clearance prediction, and MPC-style filtering ensures that constraints are enforced for the executed controls. Together, these components provide a principled high-probability safety guarantee for driving under episodic nonstationarity, while allowing the method to become less conservative as predictive accuracy improves.

\end{document}